\documentclass[final]{cvpr}

\usepackage{times}
\usepackage{epsfig}
\usepackage{graphicx}
\usepackage{amsmath}
\usepackage{amssymb}
\usepackage{subcaption}

\usepackage{bm}
\usepackage{bbm}
\usepackage{array}
\usepackage{pifont}
\usepackage{mathrsfs}
\usepackage{url}
\usepackage{multirow}
\usepackage{tablefootnote}
 
\usepackage{color}
\usepackage[ruled,linesnumbered,noend]{algorithm2e}
\SetKwInput{KwInput}{Input}                
\SetKwInput{KwOutput}{Output}

\DeclareMathOperator*{\argmin}{arg\,min} %

\usepackage[pagebackref=true,breaklinks=true,colorlinks,bookmarks=false]{hyperref}

\def\cvprPaperID{6708} 
\def\confYear{CVPR 2021}
\newcommand{\vect}[1]{\boldsymbol{#1}}
\def\vx{{\bm{x}}}
\def\vz{{\bm{z}}}
\def\vw{{\bm{w}}}
\def\vth{{\vect{\theta}}}
\def\vdelta{{\vect{\delta}}}
\newcommand{\E}{\mathbb{E}}
\newcommand{\KL}{D_{\mathrm{KL}}}

\newcommand{\RN}[1]{%
	\textup{\lowercase\expandafter{\it \romannumeral#1}}%
}

\begin{document}

\title{LiBRe: A Practical Bayesian Approach to Adversarial Detection}

\author{{Zhijie Deng}$^1$, {Xiao Yang}$^1$, {Shizhen Xu}$^2$, {Hang Su}$^1$$^\ast$, {Jun Zhu}$^1$\thanks{Corresponding author}\\
$^1$\,{\fontsize{11}{13}\selectfont Dept. of Comp. Sci. and Tech., BNRist Center, Institute for AI, Tsinghua-Bosch Joint ML Center, THBI Lab}\\
$^1$\,{\fontsize{11}{13}\selectfont Tsinghua University, Beijing, 100084, China} \;\; $^2$\,{\fontsize{11}{13}\selectfont RealAI}\\
{\tt\footnotesize \{dzj17,yangxiao19\}@mails.tsinghua.edu.cn, shizhen.xu@realai.ai, \{suhangss,dcszj\}@tsinghua.edu.cn}
}

\maketitle

\begin{abstract}
Despite their appealing flexibility, deep neural networks (DNNs) are vulnerable against adversarial examples. Various adversarial defense strategies have been proposed to resolve this problem, but they typically demonstrate restricted practicability owing to unsurmountable compromise on universality, effectiveness, or efficiency.
In this work, we propose a more practical approach, Lightweight Bayesian Refinement (LiBRe), in the spirit of leveraging Bayesian neural networks (BNNs) for adversarial detection.
Empowered by the task and attack agnostic modeling under Bayes principle, LiBRe can endow a variety of pre-trained task-dependent DNNs with the ability of defending heterogeneous adversarial attacks at a low cost. 
We develop and integrate advanced learning techniques to make LiBRe appropriate for adversarial detection.
Concretely, we build the few-layer deep ensemble variational and adopt the pre-training \& fine-tuning workflow to boost the effectiveness and efficiency of LiBRe.
We further provide a novel insight to realise adversarial detection-oriented uncertainty correction without inefficiently crafting adversarial examples during training.
Extensive empirical studies covering a wide range of scenarios verify the practicability of LiBRe.
We also conduct thorough ablation studies to evidence the superiority of our modeling and learning strategies.\footnote{Code at \url{https://github.com/thudzj/ScalableBDL}.}

\end{abstract}

\vspace{-1.7ex}
\section{Introduction}
\vspace{-0.3ex}

The blooming development of deep neural networks (DNNs) has brought great success in extensive industrial applications, such as image classification~\cite{he2016deep}, face recognition~\cite{deng2019arcface} and object detection~\cite{redmon2016you}. 
However, despite their promising expressiveness, DNNs are highly vulnerable to adversarial examples~\cite{szegedy2013intriguing,goodfellow2014explaining}, which are generated by adding human-imperceptible perturbations upon clean examples to deliberately cause misclassification, partly due to their non-linear and black-box nature. 
The threats from adversarial examples have been witnessed in a wide spectrum of practical systems~\cite{Sharif2016Accessorize,dong2019efficient}, raising an urgent requirement for advanced techniques to achieve robust and reliable decision making, especially in safety-critical scenarios~\cite{eykholt2018physical}.

Though increasing methods have been developed to tackle adversarial examples~\cite{madry2018towards,zhang2019theoretically,kannan2018adversarial,gong2017adversarial,zhang2018detecting}, they are not problemless.
On on hand, as one of the most popular adversarial defenses, adversarial training~\cite{madry2018towards,zhang2019theoretically} introduces adversarial examples into training to explicitly tailor the decision boundaries, which, yet, causes added training overheads and typically leads to degraded predictive performance on clean examples.
On the other hand, adversarial detection methods bypass the drawbacks of modifying the original DNNs by deploying a workflow to detect the adversarial examples ahead of decision making, by virtue of auxiliary classifiers~\cite{metzen2017detecting,gong2017adversarial,zhang2018detecting,carrara2018adversarial} or designed statistics~\cite{feinman2017detecting,ma2018characterizing}. 
Yet, they are usually developed for specific tasks (e.g., image classification~\cite{zhang2018detecting,lee2018simple,gong2017adversarial}) or for specific adversarial attacks~\cite{lu2018limitation}, lacking the flexibility to effectively generalize to other tasks or attacks.

\begin{figure}[t]
\centering
\begin{subfigure}[b]{0.97\columnwidth}
\centering
    \includegraphics[width=\textwidth]{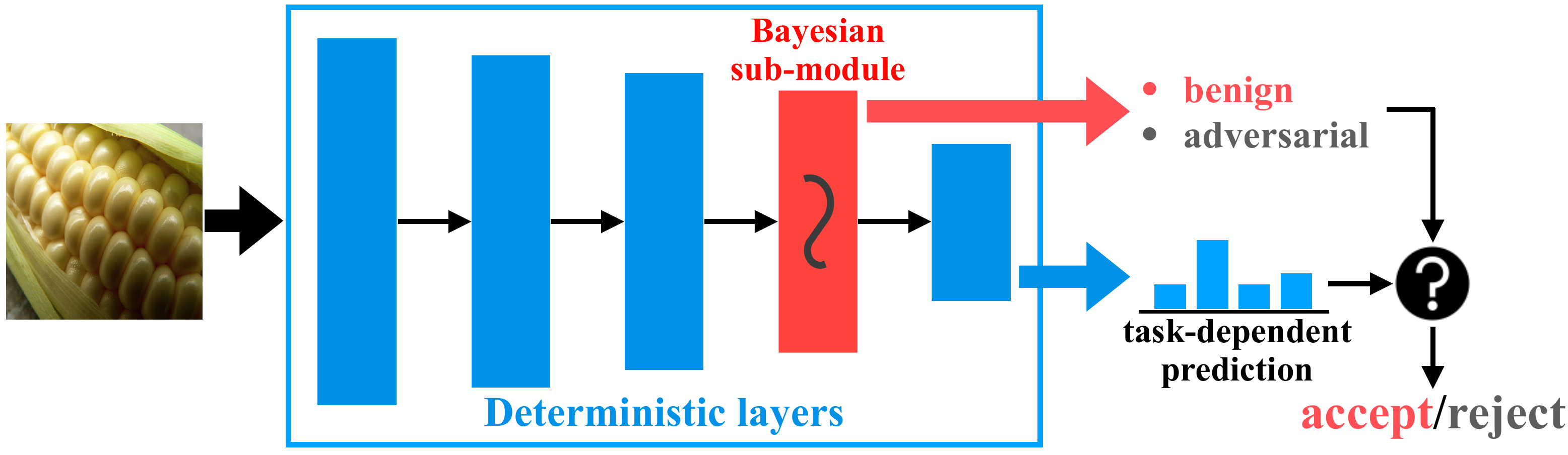}
    \vspace{-3ex}
\end{subfigure}
\caption{\small Given a pre-trained DNN, \emph{LiBRe} converts its last few layers (excluding the task-dependent output head) to be Bayesian, 
and reuses the pre-trained parameters. 
Then, \emph{LiBRe} launches several-round adversarial detection-oriented fine-tuning to render the posterior effective for prediction and meanwhile appropriate for adversarial detection. 
In the inference phase, \emph{LiBRe} estimates the predictive uncertainty and task-dependent predictions of the input concurrently, where the former is used for adversarial detection and determines the fidelity of the latter.}
\label{fig:framework}
\vspace{-2.4ex}
\end{figure}

By regarding the adversarial example as a special case of out-of-distribution (OOD) data, Bayesian neural networks (BNNs) have shown promise in adversarial detection~\cite{feinman2017detecting,louizos2017multiplicative,smith2018understanding}. 
In theory, the predictive uncertainty acquired under Bayes principle suffices for detecting heterogeneous adversarial examples in various tasks.
However, in practice, BNNs without a sharpened posterior often present systematically worse performance than their deterministic counterparts~\cite{wenzel2020good}; also relatively low-cost Bayesian inference methods frequently suffer from mode collapse and hence unreliable uncertainty~\cite{fort2019deep}. 
BNNs' requirement of more expertise for implementation and more efforts for training than DNNs further undermine their practicability.

In the work, we aim to develop a more practical adversarial detection approach by overcoming the aforementioned issues of BNNs.
We propose \emph{Lightweight Bayesian Refinement} (\emph{LiBRe}), depicted in Fig.~\ref{fig:framework}, to reach a good balance among \emph{predictive performance}, \emph{quality of uncertainty estimates} and \emph{learning efficiency}. 
Concretely, \emph{LiBRe} follows the {stochastic variational inference} pipeline~\cite{blundell2015weight}, but is empowered by two non-trivial designs: (\RN{1}) To achieve efficient learning with high-quality outcomes, we devise the \emph{Few-lAyer Deep Ensemble} (FADE) variational, which is reminiscent of Deep Ensemble~\cite{lakshminarayanan2017simple}, one of the most effective Bayesian marginalization methods~\cite{wilson2020bayesian}, and meanwhile inspired by the scalable last-layer Bayesian inference~\cite{kristiadi2020being}. 
Namely, FADE only performs \emph{deep ensemble} in the \emph{last few layers} of a model due to their crucial role for determining model behaviour, while keeps the other layers deterministic.
To encourage various ensemble candidates to capture diverse function modes, we develop a stochasticity-injected learning principle for FADE, which also benefits to reduce the gradient variance of the parameters. 
(\RN{2}) To further ease and accelerate the learning, we propose a \emph{Bayesian refinement} paradigm, where we initialize the parameters of FADE with the parameters of its pre-trained deterministic counterpart, thanks to the high alignment between FADE and point estimate.
We then perform \emph{fine-tuning} to constantly improve the FADE posterior. 

As revealed by~\cite{grosse2018limitations}, the uncertainty quantification purely acquired from Bayes principle may be unreliable for perceiving adversarial examples, thus it is indispensable to pursue an adversarial detection-oriented uncertainty correction.
For universality, we place no assumption on the adversarial examples to detect, so we cannot take the common strategy of integrating the adversarial examples crafted by specific attacks into detector training~\cite{ma2018characterizing}. 
Alternatively, we cheaply create \emph{uniformly perturbed} examples and demand high predictive uncertainty on them during Bayesian refinement to make the model be sensitive to data with any style of perturbation.
Though such a correction renders the learned posterior slightly deviated from the true Bayesian one, it can significantly boost adversarial detection performance.

The \emph{task and attack agnostic} designs enable \emph{LiBRe} to quickly and cheaply endow a pre-trained task-dependent DNN with the ability to detect various adversarial examples when facing new tasks, as testified by our empirical studies in Sec~\ref{sec:exp}.
Furthermore, \emph{LiBRe} has significantly higher inference (i.e., testing) speed than typical BNNs thanks to the adoption of \emph{lightweight variational}. 
We can achieve further speedup by exploring the potential of parallel computing, giving rise to inference speed close to the DNN in the same setting.
Extensive experiments in scenarios ranging from image classification, face recognition, to object detection confirm these claims and testify the superiority of \emph{LiBRe}.
We further perform thorough ablation studies to deeply understand the adopted modeling and learning strategies.

\vspace{-.5ex}
\section{Related Work}
\vspace{-.5ex}
Detecting adversarial examples to bypass their safety threats has attracted increasing attention recently.
Many works aim at distinguishing adversarial examples from benign ones via an auxiliary classifier applied on statistical features~\cite{gong2017adversarial,zhang2018detecting,carrara2018adversarial,cohen2020detecting,xu2017feature}. \cite{grosse2017statistical} introduces an extra class in the classifier for adversarial examples. Some recent works exploit neighboring statistics to construct more powerful detection algorithms:  \cite{lee2018simple} fits a Gaussian mixture model of the network responses, 
and resorts to the Mahalanbobis distance 
for adversarial detection in the inference phase; \cite{ma2018characterizing} introduces the more advanced local intrinsic dimensionality to describe the distance distribution and observes better results. 
{RCE~\cite{pang2018towards} is developed with the promise of leading to an enhanced distance between adversarial and normal images for kernel density~\cite{feinman2017detecting} based detection.}
However, most of the aforementioned methods are restricted in the classification scope, and the detectors trained against certain attacks may not effectively generalize to unseen attacks~\cite{lu2018limitation}.

Bayesian deep learning~\cite{graves2011practical,welling2011bayesian,blundell2015weight,balan2015bayesian,liu2016stein,kendall2017uncertainties} provides us with a more theoretically appealing way to adversarial detection.
However, though the existing BNNs manage to perceive adversarial examples~\cite{feinman2017detecting,rawat2017adversarial,smith2018understanding,louizos2017multiplicative,pawlowski2017implicit,li2017dropout}, they are typically limited in terms of training efficiency, predictive performance, \emph{etc.}, and thus cannot effectively scale up to real-world settings.
More severely, the uncertainty estimates given by the BNNs for adversarial examples are not always reliable~\cite{grosse2018limitations}, owing to the lack of particular designs for adversarial detection.
In this work, we address these issues with elaborated techniques and establish a more practical adversarial detection approach.




\vspace{-.5ex}
\section{Background}
\vspace{-.5ex}
In this section, we motivate \emph{Lightweight Bayesian Refinement} (\emph{LiBRe}) by briefly reviewing the background of adversarial defense, and then describe the general workflow of Bayesian neural networks (BNNs). 

\subsection{Adversarial Defense}
Typically, let $\mathcal{D}=\{(\vx_i, y_i)\}_{i=1}^n$ denote a collection of $n$ training samples with $\vx_i\in\mathbb{R}^d$ and $y_i \in \mathcal{Y}$ as the input data and label, respectively.
A deep neural network (DNN) parameterized by $\vw \in \mathbb{R}^p$ is frequently trained 
via \emph{maximum a posteriori} estimation (MAP):
\begin{equation}
\vspace{-.5ex}
    \label{eq:mle}
    \small
    \max_\vw \frac{1}{n}\sum_{i=1}^n \log p(y_i|\vx_i; \vw) + \frac{1}{n}\log p(\vw),
\vspace{-.5ex}
\end{equation}
where $p(y|\vx; \vw)$ refers to the predictive distribution of the DNN model.
By setting the prior $p(\vw)$ as an isotropic Gaussian, the second term amounts to the L2 (weight decay) regularizer with a tunable coefficient $\lambda$ in optimization.
Generally speaking, the adversarial example corresponding to $(\vx_i, y_i)$ against the model is defined as 
\begin{equation}
    \label{eq:adv-def}
    \small
    \vx_i^{\text{adv}} = \vx_i + \argmin_{\vdelta_i\in\mathcal{S}}\log p(y_i|\vx_i+\vdelta_i;\vw),
\end{equation}
where $\mathcal{S}=\{\vdelta: \lVert\vdelta\rVert \leq \epsilon\}$ is the valid perturbation set with $\epsilon > 0$ as the perturbation budget and $\lVert\cdot\rVert$ as some norm (e.g., ${l}_{\infty}$).
Extensive attack methods have been developed with promise to solve the above minimization problem~\cite{goodfellow2014explaining,madry2017towards,carlini2017towards,uesato2018adversarial}, based on gradients or not.


The central goal of adversarial defense is to protect the model from making undesirable decisions for the adversarial examples $\vx_i^{\text{adv}}$.
A representative line of work approaches this objective by augmenting the training data with on-the-fly generated adversarial examples and forcing the model to yield correct predictions on them~\cite{madry2018towards,zhang2019theoretically}.
But their limited training efficiency and compromising performance on clean data pose a major obstacle for real-world adoption.
As an alternative, adversarial detection methods focus on distinguishing the adversarial examples from the normal ones so as to bypass the potentially harmful outcomes of making decisions for adversarial examples~\cite{metzen2017detecting,carrara2018adversarial,ma2018characterizing}.
However, satisfactory transferability to unseen attacks and tasks beyond image classification remains elusive~\cite{lu2018limitation}.

\subsection{Bayesian Neural Networks}
In essence, the problem of distinguishing adversarial examples from benign ones can be viewed as a specialized out-of-distribution (OOD) detection problem of particular concern in safety-sensitive scenarios -- with the model trained on the clean data, we expect to identify the adversarial examples from a shifted data manifold, though the shift magnitude may be subtle and human-imperceptible. 
In this sense, we naturally introduce BNNs into the picture attributed to their principled OOD detection capacity along with the equivalent flexibility for data fitting as DNNs.

\textbf{Modeling and training.} Typically, a BNN is specified by a parameter prior $p(\vw)$ and an NN-instantiated data likelihood $p(\mathcal{D}|\vw)$.
We are interested in the parameter posterior $p(\vw|\mathcal{D})$ instead of a point estimate as in DNN.
It is known that precisely deriving the posterior is intractable owing to the high non-linearity of neural networks. 
Among the wide spectrum of approximate Bayesian inference methods, variational BNNs are particularly attractive due to their close resemblance to standard backprop~\cite{graves2011practical,blundell2015weight,louizos2016structured,sun2017learning,sun2018functional,shi2018spectral,osawa2019practical}.
Generally, in variational BNNs, we introduce a variational distribution $q(\vw|\vth)$ with parameters $\vth$ and maximize the evidence lower bound (ELBO) for learning (scaled by $1/n$):
\begin{equation}
    \label{eq:elbo}
    \small
    \max_\vth \E_{q(\vw|\vth)}\left[\frac{1}{n}\sum_{i=1}^n \log p(y_i|\vx_i; \vw)\right] - \frac{1}{n}\KL(q(\vw|\vth) \Vert p(\vw)).
\end{equation}

\textbf{Inference.} The obtained posterior $q(\vw|\vth)$\footnote{We use $q(\vw|\vth)$ equivalently with $p(\vw|\mathcal{D})$ in the following if there is no misleading.} offers us the opportunities to predict robustly.
For computational tractability, we usually estimate the \emph{posterior predictive} via: 
\begin{equation}
    \label{eq:pred}
    \small
    p(y|\vx, \mathcal{D}) = \E_{q(\vw|\vth)}\left[p(y|\vx;\vw)\right] 
    \approx \frac{1}{T}\sum_{t=1}^T p(y|\vx; \vw^{(t)}),
\end{equation}
where $\vw^{(t)} \sim q(\vw|\vth), t=1,...,T$ denote the Monte Carlo (MC) samples.
In other words, the BNN assembles the predictions yielded by all likely models to make more reliable and calibrated decisions, in stark contrast to the DNN which only cares about the most possible parameter point.

\textbf{Measure of uncertainty.} For adversarial detection, we are interested in the \emph{epistemic} uncertainty which is indicative of covariate shift. 
A superior choice of uncertainty metric is the \emph{softmax variance} given its previous success for adversarial detection in image classification~\cite{feinman2017detecting} and insightful theoretical support~\cite{smith2018understanding}.
However, the softmax output of the model may be less attractive during inference (e.g., in open-set face recognition), letting alone that not all the computer vision tasks can be formulated as pure classification problems (e.g., object detection).
To make the metric faithful and readily applicable to diverse scenarios, we concern the \emph{predictive variance of the hidden feature} $\vz$ corresponding to $\vx$, by mildly assuming the information flow inside the model as $\vx \xrightarrow[]{} \vz \xrightarrow[]{} y$. 
We utilize an unbiased variance estimator and summarize the variance of all coordinates of $\vz$ into a scalar via:
\begin{equation}
    \label{eq:mi}
    \small
    {U}(\vx) = \frac{1}{T-1}\left[\sum_{t=1}^T \lVert \vz^{(t)} \rVert^2_2 - T(\lVert \frac{1}{T}\sum_{t=1}^T \vz^{(t)} \rVert^2_2)\right],
\end{equation}
where $\vz^{(t)}$ denotes the features of $\vx$ under parameter sample $\vw^{(t)}\sim q(\vw|\vth), t=1,...,T$, with $\lVert \cdot \rVert_2$ as $\ell_2$ norm.
It is natural to simultaneously make prediction and quantify uncertainty via Eq.~(\ref{eq:pred}) and Eq.~(\ref{eq:mi}) when testing.

\section{Lightweight Bayesian Refinement}
\label{sec:method}
Despite their theoretical appealingness, BNNs are seldom adopted for real-world adversarial detection, owing to a wide range of concerns on their \emph{training efficiency}, \emph{predictive performance}, \emph{quality of uncertainty estimates}, and \emph{inference speed}.
In this section, we provide detailed and novel strategies to relieve these concerns and build the practical \emph{Lightweight Bayesian Refinement} (\emph{LiBRe}) framework.

\textbf{Variational configuration.} At the core of variational BNNs lies the configuration of the variational distribution. 
The recent surge of \emph{variational Bayes} has enabled us to leverage mean-field Gaussian~\cite{blundell2015weight}, matrix-variate Gaussian~\cite{louizos2016structured,sun2017learning}, multiplicative normalizing flows~\cite{louizos2017multiplicative} and even implicit distributions~\cite{li2017gradient,shi2018spectral} to build expressive and flexible variational distributions.
However, on one side, there is evidence to suggest that more complex variationals are commonly accompanied with less user-friendly and less scalable inference processes; 
on the other side, more popular and more approachable variationals like mean-field Gaussian, low-rank Gaussian~\cite{fort2019deep} and MC Dropout~\cite{gal2016dropout} tend to concentrate on a single mode in the function space, rendering the yielded uncertainty estimates unreliable~\cite{fort2019deep}. 

Deep Ensemble~\cite{lakshminarayanan2017simple}, a powerful alternative to BNNs, builds a set of parameter candidates $\vth=\{\vw^{(c)}\}_{c=1}^C$, which are separately trained to account for diverse function modes, and uniformly assembles their corresponding predictions for inference. 
In a probabilistic view, Deep Ensemble builds the variational 
$
    \label{eq:q}
    \small
    q(\vw|\vth) = \frac{1}{C}\sum_{c=1}^C \delta(\vw - \vw^{(c)}) 
$
with $\delta$ as the Dirac delta function.
Yet, obviously, optimizing the parameters of such a variational is computationally prohibitive~\cite{lakshminarayanan2017simple}.
Motivated by the success of last-layer Bayesian inference~\cite{kristiadi2020being}, we propose to only convert the \emph{last few layers} of the feature extraction module of a DNN, e.g., the last residual block of ResNet-50~\cite{he2016deep}, to be Bayesian layers whose parameters take the deep ensemble variational.

Formally, breaking down $\vw$ into $\vw_{b}$ and $\vw_{-b}$, which denote the parameters of the tiny Bayesian sub-module and the other parameters in the model respectively, we devise the \emph{Few-lAyer Deep Ensemble} (FADE) variational:
\begin{equation}
\vspace{-0.8ex}
    \label{eq:qs}
    \small
    q(\vw|\vth) = \frac{1}{C}\sum_{c=1}^C \delta(\vw_{b} - \vw_{b}^{(c)})\delta(\vw_{-b} - \vw_{-b}^{(0)}),
\vspace{-0.ex}
\end{equation}
where $\vth=\{\vw_{-b}^{(0)}, \vw_{b}^{(1)}, ..., \vw_{b}^{(C)}\}$.
Intuitively, FADE will strikingly ease and accelerate the learning, permitting scaling Bayesian inference up to deep architectures trivially.

\textbf{ELBO maximization.} Given the FADE variational, we develop an effective and user-friendly implementation for learning. 
Equally assuming an isotropic Gaussian prior as the MAP estimation for DNN, the second term of the ELBO in Eq.~(\ref{eq:elbo}) boils down to weight decay regularizers with coefficients ${\lambda}$ on $\vw_{-b}^{(0)}$ and $\frac{\lambda}{C}$ on $\vw_{b}^{(c)}, c=1,...,C$, which can be easily implemented inside the optimizer.\footnote{The derivation is based on relaxing the Dirac distribution as Gaussian with small variance. See Sec 3.4 of \cite{gal2015dropout} for detailed derivation insights.}
Then, we only need to explicitly deal with the first term in the ELBO.
Analytically estimating the expectation in this term is feasible but may hinder different parameter candidates from exploring diverse function modes (as they may undergo similar optimization trajectories).
Thus, we advocate maximizing a stochastic estimate of it on top of stochastic gradient ascent:
\begin{equation}
\vspace{-0.7ex}
    \small
    \max_\vth \mathcal{L} = \frac{1}{|\mathcal{B}|}\sum_{(\vx_i, y_i)\in \mathcal{B}} \log p(y_i|\vx_i; \vw_{b}^{(c)}, \vw_{-b}^{(0)}),
\vspace{-0.ex}
\end{equation}
where $\mathcal{B}$ is a stochastic mini-batch, and $c$ is drawn from $\text{unif}\{1, C\}$, i.e., the uniform distribution over $\{1, ..., C\}$.

However, intuitively, $\nabla_{\vw_{-b}^{(0)}} \mathcal{L}$ exhibits high variance across iterations due to its correlation with the varying choice of $c$, which is harmful for the convergence (see Sec~\ref{sec:abl} and \cite{kingma2015variational}). 
To disentangle such correlation, we propose to replace the batch-wise parameter sample $\vw_{b}^{(c)}$ with instance-wise ones $\vw_{b}^{(c_i)}, c_i \stackrel{i.i.d.}{\sim} \text{unif}\{1, C\}, i=1,...,|\mathcal{B}|$, which ensures $\vw_{-b}^{(0)}$ to comprehensively consider the variable behaviour of the Bayesian sub-module at per iteration.
Formally, we solve the following problem for training:
\begin{equation}
    \label{eq:loss}
    \small
    \max_\vth \mathcal{L}^* = \frac{1}{|\mathcal{B}|}\sum_{(\vx_i, y_i)\in \mathcal{B}} \log p(y_i|\vx_i; \vw_{b}^{(c_i)}, \vw_{-b}^{(0)}).
\end{equation}

Under such a learning criterion, each Bayesian parameter candidate accounts for a stochastically assigned, separate subset of $\mathcal{B}$.
Such stochasticity will be injected into the gradient ascent dynamics and serves as an implicit regularization~\cite{mandt2017stochastic}, leading $\{\vw_{b}^{(c)}\}_{c=1}^C$ to investigate diverse weight sub-spaces and ideally diverse function modes.
Compared to Deep Ensemble~\cite{lakshminarayanan2017simple} which depends on random initialization to avoid mode collapse, our approach is more theoretically motivated and more economical.

Though computing $\mathcal{L}^*$ involves the same FLOPS as computing $\mathcal{L}$, there is a barrier to make the computation compatible with modern autodiff libraries and time-saving -- \emph{de facto} computational kernels routinely process a batch given shared parameters while estimating $\mathcal{L}^*$ needs the kernels to embrace instance-specialized parameters in the Bayesian sub-module.
In spirit of parallel computing, we resort to the group convolution, batch matrix multiplication, \emph{etc.} to address this issue.
The resultant computation burden is negligibly more than the original DNN thanks to the support of powerful backends like cuDNN~\cite{chetlur2014cudnn} for these operators and the tiny size of the Bayesian sub-module.

\textbf{Adversarial example free uncertainty correction.} It is a straightforward observation that the above designs of the BNN are OOD data agnostic, leaving the ability to detect adversarial examples solely endowed by the rigorous Bayes principle.
Nevertheless, as a special category of OOD data, adversarial examples hold several special characteristics, e.g., the close resemblance to benign data and the strong offensive to the behaviour of black-box deep models, which may easily destroy the uncertainty based adversarial detection~\cite{grosse2018limitations}.
A common strategy to address this issue is to incorporate adversarial examples crafted by specific attacks into detector training~\cite{ma2018characterizing}, which, yet, is costly and may limit the learned models from generalizing to unseen attacks.
Instead, we propose an adversarial example free uncertainty correction strategy by considering a superset of the adversarial examples.
We feed uniformly perturbed training instances (which encompass all kinds of adversarial examples) into the BNN and demand relatively high predictive uncertainty on them.
Formally, with $\epsilon_{train}$ as the training perturbation budget, we perturb a mini-batch of data via 
\begin{equation}
\label{eq:perturb}
\small
    \tilde{\vx}_i = \vx_i + \vect{\delta}_i, \vect{\delta}_i \stackrel{i.i.d.}{\sim} \mathcal{U}(-\epsilon_{train}, \epsilon_{train})^d, i=1,...,|\mathcal{B}|.
\end{equation}
Then we calculate the uncertainty measure $U$ cheaply with $T=2$ MC samples, and regularize the outcome via solving the following margin loss:
\begin{equation}
\label{eq:reg}
\vspace{-0.5ex}
\small
    \max_\vth \mathcal{R} = \frac{1}{|\mathcal{B}|}\sum_{(\vx_i,y_i)\in\mathcal{B}} \text{min}(\lVert \tilde{\vz}_i^{(c_{i, 1})} - \tilde{\vz}_i^{(c_{i, 2})} \rVert_2^2, \gamma),
\vspace{-0.5ex}
\end{equation}
where $\tilde{\vz}_i^{(c_{i, j})}$ refers to the features of $\tilde{\vx}_i$ given parameter sample $\vw^{(c_{i, j})}=\{\vw_{b}^{(c_{i, j})}, \vw_{-b}^{(0)}\}$, with $c_{i, j} \stackrel{i.i.d.}{\sim} \text{unif}\{1, C\}$ and $c_{i, 1} \neq c_{i, 2}, i=1,...,|\mathcal{B}|, j=1,2$.
$\gamma$ is a tunable threshold.
Surprisingly, this regularization remarkably boosts the adversarial detection performance (see Sec~\ref{sec:abl}).

\textbf{Efficient learning by refining pre-trained DNNs.} 
Though from-scratch BNN training is feasible, a recent work demonstrate that it probably incurs worse predictive performance than a fairly trained DNN~\cite{wenzel2020good}.
Therefore, given the alignment between the posterior parameters $\vth=\{\vw_{-b}^{(0)}, \vw_{b}^{(1)}, ..., \vw_{b}^{(C)}\}$ and their DNN counterparts, we suggest to perform cost-effective Bayesian refinement upon a pre-trained DNN model, which renders our workflow more appropriate for large-scale learning.

With the pre-training DNN parameters denoted as $\vw^\dagger = \{\vw_b^\dagger, \vw_{-b}^\dagger\}$, we initialize $\vw_{-b}^{(0)}$ as $\vw_{-b}^\dagger$ and $\vw_b^{(c)}$ as $\vw_b^\dagger$ for $c=1,...,C$.
Continuing from this, we fine-tune the variational parameters to maximize $\mathcal{L}^*+\alpha\mathcal{R}$\footnote{$\alpha$ refers to a trade-off coefficient.} under weight decay regularizers with suitable coefficients to realise adversarial detection-oriented posterior inference.
The whole algorithmic procedure is presented in Algorithm~\ref{algo:1}.
Such a practical and economical refinement significantly benefits from the prevalence of open-source DNN model zoo, and is promised to maintain non-degraded predictive performance by the well-evaluated pre-training \& fine-tuning workflow.

\textbf{Inference speedup.} After learning, a wide criticism on BNNs is their requirement for longer inference time than DNNs.
This is because BNNs leverage a collection of MC samples to marginalize the posterior for prediction and uncertainty quantification, as shown in Eq.~(\ref{eq:pred}) and Eq.~(\ref{eq:mi}).
However, such a problem is desirably alleviated in our approach thanks to the adoption of the FADE variational.
The main part of the model remains deterministic, allowing us to perform only once forward propagation to reach the entry of the Bayesian sub-module.
In the Bayesian sub-module, we expect to take all the $C$ parameter candidates into account for prediction to thoroughly exploit their heterogeneous predictive behaviour, i.e., $T=C$.
Naively sequentially calculating the outcomes under each parameter candidate $\vw_b^{(c)}$ is viable, but we can achieve further speedup by unleashing the potential of parallel computing.
Take the convolution layer in the front of the Bayesian sub-module as an example (we abuse some notations here): Given a batch of features $\vx_\text{in} \in \mathbb{R}^{b \times i \times h \times w}$ and $C$ convolution kernels $\vw^{(c)} \in \mathbb{R}^{o \times i \times k \times k}, c=1, ..., C$, we first repeat $\vx_\text{in}$ at the channel dimension for $C$ times, getting $\vx'_\text{in}=\mathbb{R}^{b \times Ci \times h \times w}$, and concatenate $\{\vw^{(c)}\}_{c=1}^C$ as $\vw' \in \mathbb{R}^{Co \times i \times k \times k}$.
Then, we estimate the outcomes in parallel via group convolution: $\vx'_\text{out}=\text{conv}(\vx'_\text{in}, \vw', \text{groups}=C)$, and the outcome corresponding to $\vw^{(c)}$ is $\vx_\text{out}^{(c)}=\vx'_\text{out}[:, co-o:co, ...]$.
The cooperation between FADE variational and the above strategy makes our inference time close to that of the DNNs in the same setting (see Sec~\ref{sec:abl}), while only our approach enjoys the benefits from Bayes principle and is able to achieve robust adversarial detection. 

\begin{algorithm}[t] 
\small
\SetKwInput{KwInput}{Input} 
\DontPrintSemicolon
  \KwInput{pre-trained DNN parameters $\vw^\dagger$, 
  weight decay coefficient $\lambda$, 
  threshold $\gamma$, trade-off coefficient $\alpha$
  } 
  \SetKwFunction{FMain}{search}
  \SetKwProg{Fn}{Function}{:}{}
  Initialize $\{\vw_{b}^{(c)}\}_{c=1}^C$ and $\vw_{-b}^{(0)}$ based on $\vw^\dagger$\;
  Build optimizers $\text{opt}_b$ and $\text{opt}_{-b}$ with weight decay $\lambda/C$ and $\lambda$ for $\{\vw_{b}^{(c)}\}_{c=1}^C$ and $\vw_{-b}^{(0)}$ respectively\;

   \For{epoch = 1, 2, ..., E}{
       \For{mini-batch $\mathcal{B}=\{(\vx_i, y_i)\}_{i=1}^{|\mathcal{B}|}$ in $\mathcal{D}$}{
        Estimate the log-likelihood $\mathcal{L}^*$ via Eq.~(\ref{eq:loss}) \;
        Uniformly perturb the clean data via Eq.~(\ref{eq:perturb}) \; 
        Estimate the uncertainty penalty $\mathcal{R}$ via Eq.~(\ref{eq:reg}) \;
        Backward the gradients of $\mathcal{L}^*+\alpha\mathcal{R}$ via autodiff\;
        Perform 1-step gradient ascent with $\text{opt}_b$ \& $\text{opt}_{-b}$\;
       }
   }
\caption{\small Lightweight Bayesian Refinement} 
\label{algo:1}
\end{algorithm}
\setlength{\textfloatsep}{10pt}

\vspace{-2ex}
\section{Experiments}
\vspace{-1.ex}
\label{sec:exp}

\begin{table*}[t]

  \centering
 \footnotesize
  \begin{tabular}{c||p{17.5ex}<{\centering}p{17.5ex}<{\centering} |p{13ex}<{\centering}p{13ex}<{\centering}p{13ex}<{\centering}p{13ex}<{\centering}}
  \hline
\multirow{2}{*}{Method}& \multicolumn{2}{c|}{Prediction accuracy $\uparrow$} &\multicolumn{4}{c}{AUROC of adversarial detection under \emph{model transfer} $\uparrow$} \\
\cline{2-7}
& TOP1 & TOP5 & PGD & MIM & TIM & DIM\\
\hline
\emph{MAP} & 76.13\% &92.86\%  & - &- &- & - \\
\emph{MC dropout}~\cite{gal2016dropout}  & 74.86\% &92.33\% & 0.660 & 0.723& 0.695 & 0.605 \\
\emph{LMFVI}  & 76.06\% &92.92\%  & 0.125 & 0.200 & 0.510 & 0.018 \\
\emph{MFVI}  & 75.24\% &92.58\%  & 0.241 & 0.205 & 0.504 & 0.150 \\
\hline
\emph{LiBRe} & \textbf{76.19\%} & \textbf{92.98\%} & \textbf{1.000} & \textbf{1.000} & \textbf{0.982} & \textbf{1.000} \\
  \hline
   \end{tabular}
    \vspace{-0.25cm}
  \caption{\small Left: comparison on accuracy. Right: comparison on AUROC of adversarial detection under \emph{model transfer}. (ImageNet)}
  \label{table:imagenet-more}
  \vspace{-1ex}
\end{table*}

\begin{table*}[t]
  \centering
 \footnotesize
  \begin{tabular}{c||p{7ex}<{\centering}p{7ex}<{\centering}p{7ex}<{\centering}p{7ex}<{\centering}p{7ex}<{\centering}p{7ex}<{\centering}p{7ex}<{\centering}p{9ex}<{\centering}p{8ex}<{\centering}p{8ex}<{\centering}}
  \hline
Method & FGSM & BIM & C\&W & PGD & MIM & TIM & DIM & FGSM-$\ell_2$ & BIM-$\ell_2$ & PGD-$\ell_2$ \\
\hline
\emph{KD}~\cite{feinman2017detecting} & 0.639 & \underline{1.000} & \underline{0.999} & \underline{1.000} & \underline{1.000} & \underline{0.999} & 0.624 & 0.633 & \underline{1.000} & \underline{1.000}\\
\emph{LID}~\cite{ma2018characterizing}  & 0.846 & \underline{0.999} & \underline{0.999} & \underline{0.999} & \underline{0.997} & \underline{0.999} & 0.762 & 0.846 & \underline{0.999} & \underline{0.999}\\
\emph{MC dropout}~\cite{gal2016dropout} & 0.607 & \underline{1.000} & \underline{0.980} & \underline{1.000}& \underline{1.000} & \underline{0.999} & 0.628 & 0.577 & \underline{0.999} & \underline{0.999}\\
\emph{LMFVI} & 0.029 & \underline{0.992} & 0.738 & 0.943 & \underline{0.996} & \underline{0.997}& 0.021 & 0.251 & \underline{0.993} & 0.946\\
\emph{MFVI}& 0.102 & \underline{ 1.000} & 0.780 & \underline{0.992} & \underline{1.000} & \underline{0.999} & 0.298 & 0.358 & 0.952 & 0.935\\
\hline
\emph{LiBRe}  & \textbf{1.000} & \underline{0.984} & \underline{0.985} & \underline{0.994} & \underline{0.996} & \underline{0.994} & \textbf{1.000} & \textbf{0.995} & \underline{0.983} & \underline{0.993}\\
  \hline
   \end{tabular}
      \vspace{-0.25cm}
   \caption{\small Comparison on AUROC of adversarial detection for \emph{regular attacks} $\uparrow$. (ImageNet)}
  \label{table:imagenet}
  \vspace{-3ex}
\end{table*}

To verify if \emph{LiBRE} could quickly and economically equip the pre-trained DNNs with principled adversarial detection ability in various scenarios, we perform extensive empirical studies covering ImageNet classification~\cite{imagenet_cvpr09}, open-set face recognition~\cite{yi2014learning}, and object detection~\cite{lin2014microsoft} in this section.


\textbf{General setup.}
We fetch the pre-trained DNNs available online, and inherit all their settings for the Bayesian refinement unless otherwise stated. 
We use $C=20$ candidates for FADE across scenarios.
The FADE posterior is generally employed for the parameters of the last convolution block (e.g., the last residual block for ImageNet and face tasks or the feature output heads for object detection).
We take the immediate output of the Bayesian sub-module as $\vz$ for estimating \emph{feature variance} uncertainty. 

\textbf{Attacks.} We adopt some popular attacks to craft adversarial examples under $\ell_{2}$ and $\ell_{\infty}$ threat models, including fast gradient sign method (FGSM)~\cite{goodfellow2014explaining}, basic iterative method (BIM)~\cite{kurakin2016adversarial}, projected gradient descent method (PGD)~\cite{madry2017towards}, momentum iterative method (MIM)~\cite{Dong2017}, Carlini $\&$ Wagner’s method (C$\&$W)~\cite{carlini2017towards}, diverse
inputs method (DIM)~\cite{xie2019improving}, and translation-invariant method (TIM)~\cite{dong2019evading}.
We set the perturbation budget as $\epsilon=$16/255.
We set step size as 1/255 and the number of steps as 20 for all the iterative methods.
When attacking BNNs, the minimization goal in Eq.~(\ref{eq:adv-def}) refers to the \emph{posterior predictive} in Eq.~(\ref{eq:pred}) with $T=20$.
More details are deferred to Appendix.


\textbf{Baselines.} 
Given the fact that many of the recent adversarial detection methods focus on specific tasks or attacks and hence can hardly be effectively extended to the challenging settings considered in this paper (e.g., attacks under model transfer, object detection), we mainly compare \emph{LiBRe} to baselines implemented by ourselves, including 1) the fine-tuning start point \emph{MAP}; 2) two standard adversarial detection approaches  \emph{KD}~\cite{feinman2017detecting} and \emph{LID}~\cite{ma2018characterizing}, which both work on the extracted features by \emph{MAP}; 
3) three popular BNN baselines \emph{MC dropout}~\cite{gal2016dropout}, \emph{MFVI}~\cite{blundell2015weight}, and \emph{LMFVI}.
\emph{MC dropout} trains dropout networks from scratch and enables dropout during inference.
\emph{MFVI} is trained by the typical mean-field variational inference, and \emph{LMFVI} is a \emph{lightweight} variant of it with only the last few layers converted to be Bayesian  (similar to \emph{LiBRe}).
\emph{MFVI} and \emph{LMFVI} work in a Bayesian refinement manner in analogy to \emph{LiBRe} for fair comparison.
\emph{MC dropout}, \emph{MFVI}, and \emph{LMFVI} are all trained without uncertainty calibration $\mathcal{R}$ and take the \emph{feature variance} as the measure of uncertainty.

\textbf{Metric. } The adversarial detection is essentially a binary classification, so we report the area under the receiver operating characteristic (AUROC) based on the raw predictive uncertainty (for \emph{MFVI}, \emph{LMFVI}, \emph{MC dropout}, and \emph{LiBRe}), or the output of an extra detector (for \emph{KD} and \emph{LID}).

\begin{figure*}[t]
\centering
\begin{subfigure}[b]{0.89\textwidth}
\centering
    \includegraphics[width=\textwidth]{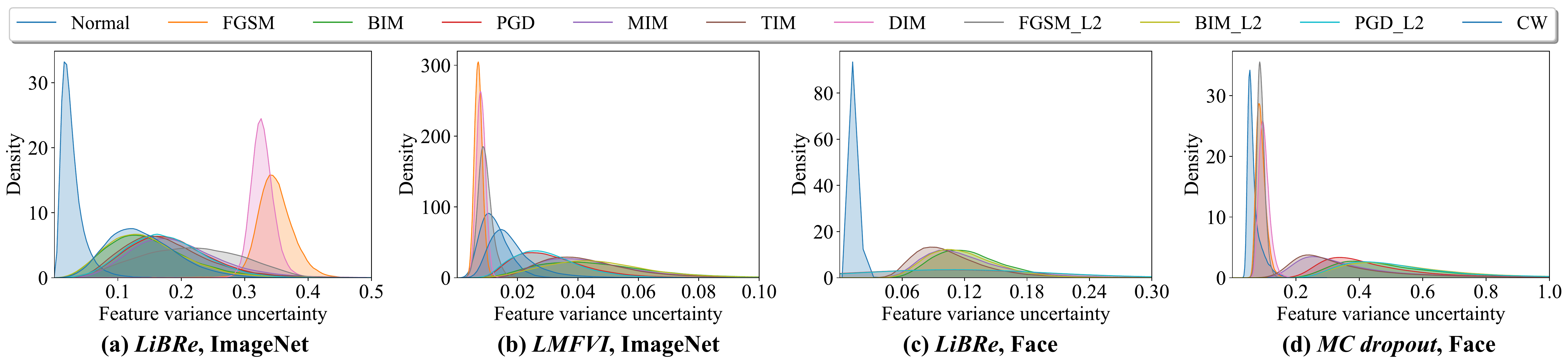}
    \vspace{-1.5ex}
\end{subfigure}
\vspace{-3.5ex}
\caption{\small The histograms for the \emph{feature variance} uncertainty of normal and adversarial examples given by \emph{LiBRe} or the baselines.}
\label{fig:density}
\vspace{-1ex}
\end{figure*}

\begin{table*}[t]
  
  \centering
 \footnotesize
  \begin{tabular}{c||p{6ex}<{\centering}p{6ex}<{\centering}p{6ex}<{\centering}p{6ex}<{\centering}|p{6ex}<{\centering}p{6ex}<{\centering}p{6ex}<{\centering}p{6ex}<{\centering}|p{6ex}<{\centering}p{6ex}<{\centering}p{6ex}<{\centering}p{6ex}<{\centering}}
  \hline
\multirow{2}{*}{Method}& \multicolumn{4}{c|}{Softmax} & \multicolumn{4}{c|}{CosFace} & \multicolumn{4}{c}{ArcFace} \\
\cline{2-13}
 & \emph{MAP} & \emph{MCD} & \emph{LMFVI} & \emph{LiBRe}& \emph{MAP} & \emph{MCD} & \emph{LMFVI} & \emph{LiBRe}& \emph{MAP} & \emph{MCD} & \emph{LMFVI} & \emph{LiBRe}\\
\hline
\emph{VGGFace2} & \textbf{0.9256} & 0.9254 & 0.9198 & 0.9246& 0.9370& 0.9370 & 0.9360  & {\textcolor{blue}{\textbf{0.9376}}}& 0.9356& 0.9334 & \textbf{0.9358}  & 0.9348 \\
\emph{LFW} & \textbf{0.9913}& 0.9898 & 0.9912  & 0.9892 & 0.9930& 0.9932 & 0.9920  & \textbf{0.9935} & 0.9933& 0.9930 & 0.9933  & \textcolor{blue}{\textbf{0.9943}}  \\
\emph{CPLFW} & 0.8630& \textbf{0.8638} & 0.8610  & 0.8598 & 0.8915 & 0.8890 & \textcolor{blue}{\textbf{0.8925}} & 0.8910 & 0.8808& 0.8803 & 0.8833  & \textbf{0.8837}  \\
\emph{CALFW} & 0.9107& 0.9110 & 0.9087  & \textbf{0.9120} &0.9327& 0.9345 & 0.9333  & \textcolor{blue}{\textbf{0.9352}} & 0.9292& \textbf{0.9300} & 0.9250  & 0.9283  \\
\emph{AgedDB-30} & \textbf{0.9177}& 0.9170 & 0.9128 & 0.9167 & \textcolor{blue}{\textbf{0.9435}}& 0.9422 & 0.9387  & 0.9433 & 0.9327& 0.9317 & \textbf{0.9337}  & \textbf{0.9337} \\
\emph{CFP-FP} & 0.9523& \textbf{0.9543} & 0.9480  & 0.9489 & 0.9564 & 0.9567 & 0.9583& \textcolor{blue}{\textbf{0.9597}} & \textbf{0.9587}& 0.9586 & 0.9554  & 0.9573 \\
\emph{CFP-FF} & 0.9873& {0.9870} & \textbf{0.9874}  & \textbf{0.9874} & \textcolor{blue}{\textbf{0.9927}} & 0.9926 & 0.9916 & \textcolor{blue}{\textbf{0.9927}} & 0.9914& 0.9910 & 0.9911  & \textbf{0.9921} \\
  \hline
   \end{tabular}
      \vspace{-0.3cm}
   \caption{\small Accuracy comparison on face recognition $\uparrow$. \emph{MCD} is short for \emph{MC dropout}. \textbf{Bold} refers to the best results under specific loss function. \textcolor{blue}{\textbf{Blue bold}} refers to the overall best results.}
  \label{table:face-more}
  \vspace{-1ex}
\end{table*}

\begin{table*}[h!]
  \centering
 \footnotesize
  \begin{tabular}{c||p{11ex}<{\centering}p{8.5ex}<{\centering}p{8.5ex}<{\centering}|p{11ex}<{\centering}p{8.5ex}<{\centering}p{8.5ex}<{\centering}|p{11ex}<{\centering}p{8.5ex}<{\centering}p{8.5ex}<{\centering}}
  \hline
\multirow{2}{*}{Attack}& \multicolumn{3}{c|}{Softmax} & \multicolumn{3}{c|}{CosFace} & \multicolumn{3}{c}{ArcFace} \\
\cline{2-10}
 & \emph{MC dropout} & \emph{LMFVI} & \emph{LiBRe} & \emph{MC dropout} & \emph{LMFVI} & \emph{LiBRe} & \emph{MC dropout} & \emph{LMFVI} & \emph{LiBRe}\\
\hline
FGSM &  0.866 & 0.155 &  \textbf{1.000}&  0.889 & 0.001 &  \textbf{1.000}&  0.794 & 0.001 &  \textbf{1.000} \\
BIM &  1.000 & 1.000  &  0.999&  1.000 &1.000  & 0.999&  1.000 & 1.000 &  1.000 \\
PGD &  1.000 & 0.992 &  0.999&  1.000 & 0.998 &  0.998&  1.000 & 0.990 &  1.000 \\
MIM &  1.000 & 1.000  & 0.999&  1.000& 1.000 &  0.999&  1.000 & 1.000 &  1.000 \\
TIM &  1.000 & 1.000&  0.999&  1.000 & 1.000 &  0.998&  1.000 & 1.000 &  1.000 \\
DIM &  0.910 & 0.025  &  \textbf{1.000}&  0.850 & 0.000 &  \textbf{1.000}&  0.746& 0.000 &  \textbf{1.000} \\
FGSM-$\ell_2$ &  0.860 & 0.659&  \textbf{1.000}&   0.825 &0.014  & \textbf{0.999}&  0.660 & 0.002 & \textbf{0.999} \\
BIM-$\ell_2$ &  1.000 & 1.000 & 0.999&  1.000 &1.000  & 1.000&  1.000 & 1.000 & 1.000 \\
PGD-$\ell_2$ &  1.000 & 0.996 & 0.999&  1.000 & 0.999  & 1.000&  1.000 & 0.994 & 1.000 \\

  \hline
   \end{tabular}
    \vspace{-0.3cm}
   \caption{\small Comparison on adversarial detection AUROC $\uparrow$. We report the averaged AUROC over the verification datasets. (face recognition)}
  \label{table:face}
  \vspace{-4ex}
\end{table*}

\subsection{ImageNet Classification}
\label{sec:imagenet}
We firstly check the adversarial detection effectiveness of \emph{LiBRe} on ImageNet.
We utilize the ResNet-50~\cite{he2016deep} architecture with weight decay coefficient $\lambda=10^{-4}$, and set the uncertainty threshold $\gamma$ as 0.5 according to the observation that the normal samples usually have $<0.5$ \emph{feature variance} uncertainty.
We set $\alpha=1$ without tuning.
We uniformly sample a training perturbation budget $\epsilon_{train} \in [\frac{\epsilon}{2}, 2\epsilon]$ at per iteration. 
We perform fine-tuning for $E=6$ epochs with learning rate of $\{\vw_{b}^{(c)}\}_{c=1}^C$ annealing from $10^{-3}$ to $10^{-4}$ with a cosine schedule and that of $\vw_{-b}^{(0)}$ fixed as $10^{-4}$.

To defend \emph{regular attacks}, \emph{KD} and \emph{LID} require to train a separate detector for every attack under the supervision of the adversarial examples from that attack. Thus, to show the \emph{best} performance of \emph{KD} and \emph{LID}, we test the trained detectors only on their corresponding adversarial examples. 
By contrast, \emph{LiBRE}, \emph{MC dropout}, \emph{LMFVI}, and \emph{MFVI} do not rely on specific attacks for training, thus have the potential to detect any (unseen) attack, which is more flexible yet more challenging. 
With that said, they can be trivially applied to detect the adversarial examples under \emph{model transfer}, which are crafted against a surrogate ResNet-152 DNN but are used to attack the trained models, to further assess the generalization ability of these defences.


The results are presented in Table~\ref{table:imagenet-more} and Table~\ref{table:imagenet}.
We also illustrate the uncertainty of normal and adversarial examples assigned by \emph{LiBRe} and a baseline in Fig.~\ref{fig:density}.
It is an immediate observation that \emph{LiBRe} preserves non-degraded prediction accuracy compared to its refinement start point \emph{MAP}, and meanwhile demonstrates \emph{near-perfect} capacity of detecting adversarial examples. 
The superiority of \emph{LiBRe} is especially apparent under the more difficult \emph{model transfer} paradigm. 
The results in Fig.~\ref{fig:density} further testify the ability of \emph{LiBRe} to assign higher uncertainty for adversarial examples to distinguish them from the normal ones.
Although \emph{KD} and the golden standard, \emph{LID}, obtain full knowledge of the models and the attacks, we can still see evident margins between their \emph{worst-case}\footnote{The worst case is of much more concern than the average for assessing robustness.} results and that of \emph{LiBRe}. 


The uncertainty-based detection baselines \emph{MC dropout}, \emph{LMFVI}, and \emph{MFVI} 
are substantially outperformed by \emph{LiBRe} when considering the worst case.
It is noteworthy that \emph{MC dropout} is slightly better than \emph{LMFVI} and \emph{MFVI} for adversarial detection, despite with worse accuracy.
We also find that the performance of \emph{LMFVI} is matched with that of \emph{MFVI}, supporting the proposed \emph{lightweight variational} notion.
Thus we use \emph{LMFVI} as a major baseline in face recognition instead of \emph{MFVI} due to its efficiency.

\vspace{-0.1cm}
\subsection{Face Recognition}
\vspace{-0.1cm}
\label{sec:face}

In this section, we concern the more realistic open-set face recognition on CASIA-WebFace~\cite{yi2014learning}. 
We adopt the IResNet-50 architecture~\cite{deng2019arcface} and try three task-dependent loss: Softmax, CosFace~\cite{wang2018cosface}, and ArcFace~\cite{deng2019arcface}. 
We follow the default hyper-parameter settings of \cite{wang2018cosface,deng2019arcface} and set $\lambda$ as $5\times 10^{-4}$.
We tune some crucial hyper-parameters according to a \emph{held-out} validation set and set $\gamma=1$, $\alpha=100$, and $E=4$.
We uniformly sample $\epsilon_{train} \in [\epsilon, 2\epsilon]$ at per iteration. We adopt the same optimizer settings as on ImageNet.
We perform comprehensive evaluation on face verification datasets
including LFW~\cite{huang2008labeled}, CPLFW~\cite{zheng2018cross}, CALFW~\cite{zheng2017cross}, CFP~\cite{sengupta2016frontal}, VGGFace2~\cite{cao2018vggface2}, and AgeDB-30~\cite{moschoglou2017agedb}.

We provide the comparison results in Table~\ref{table:face-more}, Table~\ref{table:face}, and sub-figure (c) and (d) of Fig.~\ref{fig:density}.
As expected, \emph{LiBRe} frequently yields non-degraded recognition accuracy compared to \emph{MAP}. 
Though the major goal of \emph{LiBRe} is not to boost the task-dependent performance of the pre-trained DNNs, to our surprise, \emph{LiBRe} demonstrates dominant performance under the {CosFace} loss function.
Regarding the quality of adversarial detection, \emph{LiBRe} also bypasses the competitive baselines, especially in the worst case.
These results prove the universality and practicability of \emph{LiBRe}.

\vspace{-0.1cm}
\subsection{Object Detection on COCO}
\vspace{-0.1cm}
Then, we move to a more challenging task -- object detection on COCO~\cite{lin2014microsoft}. 
Attacking and defending in object detection are more complicated and harder than in image classification~\cite{xie2017adversarial}. 
Thus, rare of the previous works have generalized their methodology into this scenario.
By contrast, the \emph{task agnostic} designs in \emph{LiBRe} make it readily applicable to object detection without compromising effectiveness.
Here, we launch experiments to identify this.

We take the state-of-the-art YOLOV5~\cite{yolov5} to perform experiments on COCO.
In detail, we setup the experiments with $\lambda=5\times 10^{-4}$, $\gamma=0.02$, and $\alpha=0.02$.
The other settings are aligned with those on face recognition.

\textbf{Multi-objective attack}. Distinct from the ordinary classifiers, the object detector exports the locations of objects along with their classification results. 
Thus, an adversary needs to perform multi-objective attack to either make the detected objects wrongly classified or render the objects of interest undetectable. 
Specifically, we craft adversarial examples by maximizing a unified loss of the two factors derived from~\cite{yolov5} w.r.t. the input image, which enables us to reuse the well developed FGSM, BIM, PGD, MIM, \emph{etc}.

Table~\ref{table:coco} exhibits the results.
As expected, \emph{LiBRe} shows satisfactory performance for detecting the four kinds of adversarial examples, verifying the universality of the Bayes principle based adversarial detection mechanism.

\begin{table}[t]
  
  \centering
 \footnotesize
  \begin{tabular}{c||p{7ex}<{\centering}p{11ex}<{\centering}|p{4ex}<{\centering}p{4ex}<{\centering}p{4ex}<{\centering}p{4ex}<{\centering}}
  \hline
\multirow{2}{*}{Method}& \multicolumn{2}{c|}{Object detection} & \multicolumn{4}{c}{Adversarial detection} \\
\cline{2-7}
& {mAP@.5} & {mAP@.5:.95} & FGSM & BIM & PGD & MIM\\
\hline
\emph{MAP} & 0.559 & 0.357 & - & - & - & -\\
\emph{LiBRe}  & 0.545 & 0.344 & 0.957 & 0.936 &  0.972 & 0.966 \\
  \hline
   \end{tabular}
    \vspace{-0.28cm}
   \caption{\small Results on object detection. (COCO)}
  \label{table:coco}
  \vspace{-1.ex}
\vspace{-0.05cm}
\end{table}

\vspace{-0.03cm}
\subsection{Ablation Study}
\vspace{-0.07cm}
\label{sec:abl}

\textbf{Comparison on uncertainty measure.} As argued, the \emph{feature variance} uncertainty is more generic than the widely used \emph{softmax variance}. 
But, do they have matched effectiveness for adversarial detection? 
Here we answer this question.
We estimate the AUROC of detecting various adversarial examples based on \emph{softmax variance} uncertainty and list the results in row 2-3 of Table~\ref{table:abl}.
Notably, the \emph{softmax variance} brings much worse detection performance than \emph{feature variance}.
We attribute this to that the transformations to produce softmax output aggressively prune the information uncorrelated with the task-dependent target, but such information is crucial for qualifying uncertainty.


\textbf{Effectiveness of $\mathcal{R}$. }
Another question of interest is whether the compromising adversarial detection performance of \emph{LMFVI} and \emph{MFVI} stems from the naive training without uncertainty regularization $\mathcal{R}$.
For an answer, we train two variants of \emph{LMFVI} and \emph{MFVI} which incorporate $\mathcal{R}$ into the training like \emph{LiBRe}.
Their results are offered in row 4-5 of Table~\ref{table:abl}.
These results reflect that training under $\mathcal{R}$ will indeed significantly boost the adversarial detection performance.
Yet, the two variants are still not as good as \emph{LiBRe}, implying the supremacy of FADE.



\textbf{Effectiveness of $\mathcal{L}^*$. } We then look at another key design of \emph{LiBRe} -- optimizing $\mathcal{L}^*$, the instance-wise stochastic estimation of the expected log-likelihood (the first term of the ELBO), rather than $\mathcal{L}$.
To deliver a quantitative analysis, we train \emph{LiBRe} by optimizing $\mathcal{L}$ and estimate the adversarial detection quality of the learned model, obtaining the results presented in row 6 of Table~\ref{table:abl}.
The obviously worse results than original \emph{LiBRe} substantiate our concerns on $\mathcal{L}$ in Sec~\ref{sec:method}.

\begin{table}[t]
  \centering
 \footnotesize
  \begin{tabular}{c|c||p{5ex}<{\centering}p{5ex}<{\centering}p{5ex}<{\centering}p{5ex}<{\centering}p{5ex}<{\centering}}
  \hline
Ablation & Method & FGSM & C\&W & PGD & DIM\\
\hline
\multirow{2}{*}{w/ SV} & \emph{MC dropout} & 0.759 & 0.013 & 0.049& 0.752\\
 & \emph{LiBRe} & 0.708 & 0.107 & 0.361& 0.650\\
\hline
\multirow{2}{*}{w/ UR} & \emph{LMFVI} & 0.990 & 0.921 & 0.980 & 0.989\\
 & \emph{MFVI} & 0.986 & 0.943 & 0.999& 0.992\\
\hline
w/ $\mathcal{L}$ & \emph{LiBRe} & 0.433 & 0.820 & 0.887  & 0.247 \\ 
  \hline
   \end{tabular}
      \vspace{-0.25cm}
   \caption{\small AUROC comparison for ablation study. As a reference, the results of \emph{LiBRe} are 1.000, 0.985, 0.994, and 1.000, respectively. SV refers to using \emph{softmax variance} as uncertainty measure. UR refers to training under the uncertainty regularization $\mathcal{R}$. $\mathcal{L}$ refers to using batch-wise MC estimation in training. (ImageNet)}
  \label{table:abl}
\end{table}

\begin{figure}[t]
\centering
\begin{subfigure}[b]{0.97\columnwidth}
\centering
    \includegraphics[width=\textwidth]{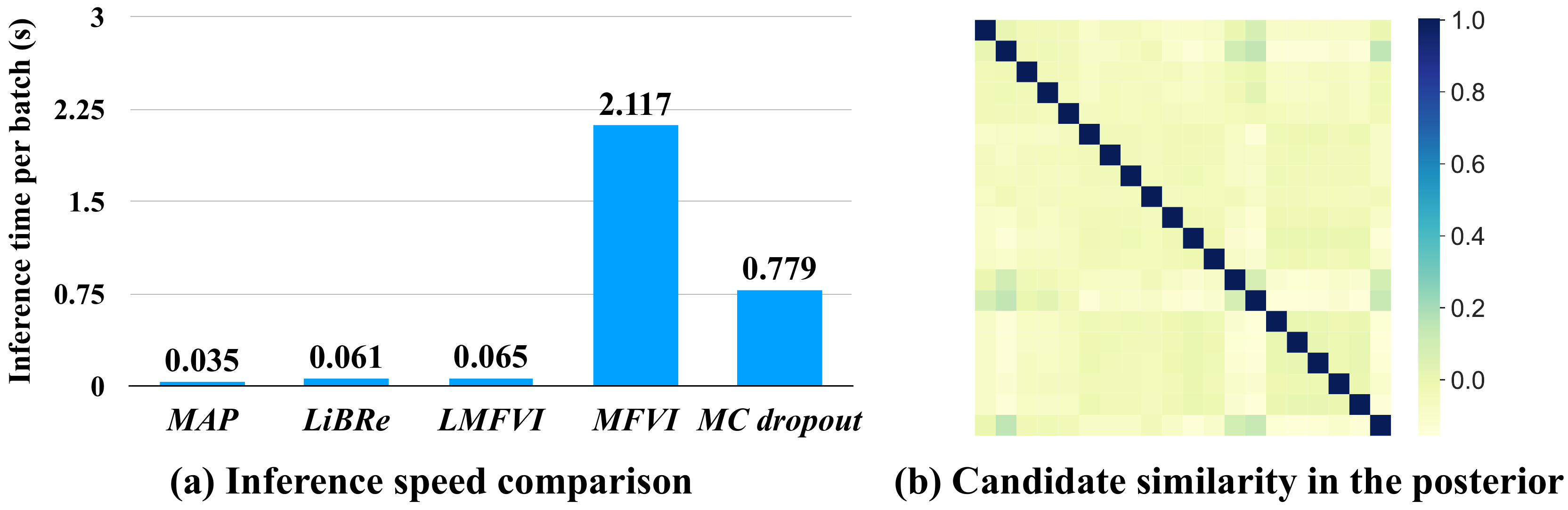}
    \vspace{-2.ex}
\end{subfigure}
\vspace{-2.2ex}
\caption{\small Left: the time for estimating the \emph{posterior predictive} of a mini-batch of 32 ImageNet instances with $T=20$ MC samples on one RTX 2080-Ti GPU (\emph{MAP} performs deterministic inference without MC estimation). Right: the similarity between the candidates in the learned FADE posterior.}
\label{fig:abl-more}
\end{figure}

\textbf{Inference speed. } We compare the inference speed of \emph{LiBRe} to the baselines in sub-figure (a) of Fig.~\ref{fig:abl-more}.
\emph{LiBRe} and \emph{LMFVI} are orders of magnitude faster than the other two BNNs.
\emph{LiBRe} is only slightly slower than \emph{MAP}, but can yield uncertainty estimates for adversarial detection.

\textbf{A visualization of the posterior.}
To verify the claim that our learning strategies lead to posteriors without mode collapse, we reduce the dimension of the candidates in the learned FADE posterior via PCA and then compute the cosine similarity between them.
Sub-figure (b) of Fig.~\ref{fig:abl-more} depicts the results, which signify the candidate diversity.

\section{Conclusion}
In this work, we propose a practical Bayesian approach to supplement the pre-trained task-dependent DNNs with the ability of adversarial detection at a low cost.
The developed strategies enhance the efficiency and the quality of adversarial detection without compromising predictive performance.
Extensive experiments validate the practicability of the proposed method. 
For future work, we can develop a parameter-sharing variant of {FADE} for higher efficiency, apply \emph{LiBRe} to DeepFake detection, etc.

\section*{Acknowledgements}
{
This work was supported by the National Key Research and Development Program of China (No.2020AAA0104304, No. 2017YFA0700904), NSFC Projects (Nos. 61620106010, 62076147, U19A2081, U19B2034, U1811461),
Beijing Academy of Artificial Intelligence (BAAI), Tsinghua-Huawei Joint Research Program, a
grant from Tsinghua Institute for Guo Qiang, Tiangong Institute for Intelligent Computing, and the
NVIDIA NVAIL Program with GPU/DGX Acceleration.
}

\clearpage
{\small
\bibliographystyle{ieee_fullname}
\bibliography{egbib}

\begin{thebibliography}{10}\itemsep=-1pt

\bibitem{balan2015bayesian}
Anoop~Korattikara Balan, Vivek Rathod, Kevin~P Murphy, and Max Welling.
\newblock {B}ayesian dark knowledge.
\newblock In {\em Advances in Neural Information Processing Systems}, pages
  3438--3446, 2015.

\bibitem{blundell2015weight}
Charles Blundell, Julien Cornebise, Koray Kavukcuoglu, and Daan Wierstra.
\newblock Weight uncertainty in neural network.
\newblock In {\em International Conference on Machine Learning}, pages
  1613--1622, 2015.

\bibitem{cao2018vggface2}
Qiong Cao, Li Shen, Weidi Xie, Omkar~M Parkhi, and Andrew Zisserman.
\newblock Vggface2: A dataset for recognising faces across pose and age.
\newblock In {\em 2018 13th IEEE International Conference on Automatic Face \&
  Gesture Recognition (FG 2018)}, pages 67--74. IEEE, 2018.

\bibitem{carlini2017towards}
Nicholas Carlini and David Wagner.
\newblock Towards evaluating the robustness of neural networks.
\newblock In {\em IEEE Symposium on Security and Privacy}, 2017.

\bibitem{carrara2018adversarial}
Fabio Carrara, Rudy Becarelli, Roberto Caldelli, Fabrizio Falchi, and Giuseppe
  Amato.
\newblock Adversarial examples detection in features distance spaces.
\newblock In {\em Proceedings of the European Conference on Computer Vision
  (ECCV)}, 2018.

\bibitem{chetlur2014cudnn}
Sharan Chetlur, Cliff Woolley, Philippe Vandermersch, Jonathan Cohen, John
  Tran, Bryan Catanzaro, and Evan Shelhamer.
\newblock cudnn: Efficient primitives for deep learning.
\newblock {\em arXiv preprint arXiv:1410.0759}, 2014.

\bibitem{cohen2020detecting}
Gilad Cohen, Guillermo Sapiro, and Raja Giryes.
\newblock Detecting adversarial samples using influence functions and nearest
  neighbors.
\newblock In {\em Conference on Computer Vision and Pattern Recognition
  (CVPR)}, 2020.

\bibitem{imagenet_cvpr09}
J. Deng, W. Dong, R. Socher, L.-J. Li, K. Li, and L. Fei-Fei.
\newblock {ImageNet: A Large-Scale Hierarchical Image Database}.
\newblock In {\em CVPR09}, 2009.

\bibitem{deng2019arcface}
Jiankang Deng, Jia Guo, Niannan Xue, and Stefanos Zafeiriou.
\newblock Arcface: Additive angular margin loss for deep face recognition.
\newblock In {\em Proceedings of the IEEE Conference on Computer Vision and
  Pattern Recognition}, pages 4690--4699, 2019.

\bibitem{Dong2017}
Yinpeng Dong, Fangzhou Liao, Tianyu Pang, Hang Su, Jun Zhu, Xiaolin Hu, and
  Jianguo Li.
\newblock Boosting adversarial attacks with momentum.
\newblock In {\em Proceedings of the IEEE Conference on Computer Vision and
  Pattern Recognition (CVPR)}, 2018.

\bibitem{dong2019evading}
Yinpeng Dong, Tianyu Pang, Hang Su, and Jun Zhu.
\newblock Evading defenses to transferable adversarial examples by
  translation-invariant attacks.
\newblock In {\em Proceedings of the IEEE Conference on Computer Vision and
  Pattern Recognition (CVPR)}, 2019.

\bibitem{dong2019efficient}
Yinpeng Dong, Hang Su, Baoyuan Wu, Zhifeng Li, Wei Liu, Tong Zhang, and Jun
  Zhu.
\newblock Efficient decision-based black-box adversarial attacks on face
  recognition.
\newblock In {\em Proceedings of the IEEE Conference on Computer Vision and
  Pattern Recognition (CVPR)}, 2019.

\bibitem{eykholt2018physical}
Kevin Eykholt, Ivan Evtimov, Earlence Fernandes, Bo Li, Amir Rahmati, Florian
  Tramer, Atul Prakash, Tadayoshi Kohno, and Dawn Song.
\newblock Physical adversarial examples for object detectors.
\newblock {\em arXiv preprint arXiv:1807.07769}, 2018.

\bibitem{feinman2017detecting}
Reuben Feinman, Ryan~R Curtin, Saurabh Shintre, and Andrew~B Gardner.
\newblock Detecting adversarial samples from artifacts.
\newblock {\em arXiv preprint arXiv:1703.00410}, 2017.

\bibitem{fort2019deep}
Stanislav Fort, Huiyi Hu, and Balaji Lakshminarayanan.
\newblock Deep ensembles: A loss landscape perspective.
\newblock {\em arXiv preprint arXiv:1912.02757}, 2019.

\bibitem{gal2015dropout}
Yarin Gal and Zoubin Ghahramani.
\newblock Dropout as a bayesian approximation: appendix.
\newblock {\em arXiv preprint arXiv:1506.02157}, 420, 2015.

\bibitem{gal2016dropout}
Yarin Gal and Zoubin Ghahramani.
\newblock Dropout as a {B}ayesian approximation: Representing model uncertainty
  in deep learning.
\newblock In {\em International Conference on Machine Learning}, pages
  1050--1059, 2016.

\bibitem{gong2017adversarial}
Zhitao Gong, Wenlu Wang, and Wei-Shinn Ku.
\newblock Adversarial and clean data are not twins.
\newblock {\em arXiv preprint arXiv:1704.04960}, 2017.

\bibitem{goodfellow2014explaining}
Ian~J Goodfellow, Jonathon Shlens, and Christian Szegedy.
\newblock Explaining and harnessing adversarial examples.
\newblock {\em arXiv preprint arXiv:1412.6572}, 2014.

\bibitem{graves2011practical}
Alex Graves.
\newblock Practical variational inference for neural networks.
\newblock In {\em Advances in Neural Information Processing Systems}, pages
  2348--2356, 2011.

\bibitem{grosse2017statistical}
Kathrin Grosse, Praveen Manoharan, Nicolas Papernot, Michael Backes, and
  Patrick McDaniel.
\newblock On the (statistical) detection of adversarial examples.
\newblock {\em arXiv preprint arXiv:1702.06280}, 2017.

\bibitem{grosse2018limitations}
Kathrin Grosse, David Pfaff, Michael~Thomas Smith, and Michael Backes.
\newblock The limitations of model uncertainty in adversarial settings.
\newblock {\em arXiv preprint arXiv:1812.02606}, 2018.

\bibitem{he2016deep}
Kaiming He, Xiangyu Zhang, Shaoqing Ren, and Jian Sun.
\newblock Deep residual learning for image recognition.
\newblock In {\em Proceedings of the IEEE Conference on Computer Vision and
  Pattern Recognition}, pages 770--778, 2016.

\bibitem{huang2008labeled}
Gary~B Huang, Marwan Mattar, Tamara Berg, and Eric Learned-Miller.
\newblock Labeled faces in the wild: A database forstudying face recognition in
  unconstrained environments.
\newblock In {\em Technical report}, 2007.

\bibitem{kannan2018adversarial}
Harini Kannan, Alexey Kurakin, and Ian Goodfellow.
\newblock Adversarial logit pairing.
\newblock {\em arXiv preprint arXiv:1803.06373}, 2018.

\bibitem{kendall2017uncertainties}
Alex Kendall and Yarin Gal.
\newblock What uncertainties do we need in {B}ayesian deep learning for
  computer vision?
\newblock In {\em Advances in Neural Information Processing Systems}, pages
  5574--5584, 2017.

\bibitem{kingma2014adam}
Diederik~P Kingma and Jimmy Ba.
\newblock Adam: A method for stochastic optimization.
\newblock {\em arXiv preprint arXiv:1412.6980}, 2014.

\bibitem{kingma2015variational}
Durk~P Kingma, Tim Salimans, and Max Welling.
\newblock Variational dropout and the local reparameterization trick.
\newblock In {\em Advances in Neural Information Processing Systems}, pages
  2575--2583, 2015.

\bibitem{kristiadi2020being}
Agustinus Kristiadi, Matthias Hein, and Philipp Hennig.
\newblock Being bayesian, even just a bit, fixes overconfidence in relu
  networks.
\newblock {\em arXiv preprint arXiv:2002.10118}, 2020.

\bibitem{kurakin2016adversarial}
Alexey Kurakin, Ian Goodfellow, and Samy Bengio.
\newblock Adversarial examples in the physical world.
\newblock {\em arXiv preprint arXiv:1607.02533}, 2016.

\bibitem{lakshminarayanan2017simple}
Balaji Lakshminarayanan, Alexander Pritzel, and Charles Blundell.
\newblock Simple and scalable predictive uncertainty estimation using deep
  ensembles.
\newblock In {\em Advances in Neural Information Processing Systems}, pages
  6402--6413, 2017.

\bibitem{lee2018simple}
Kimin Lee, Kibok Lee, Honglak Lee, and Jinwoo Shin.
\newblock A simple unified framework for detecting out-of-distribution samples
  and adversarial attacks.
\newblock In {\em Advances in Neural Information Processing Systems (NeurIPS)},
  2018.

\bibitem{li2017dropout}
Yingzhen Li and Yarin Gal.
\newblock Dropout inference in bayesian neural networks with alpha-divergences.
\newblock {\em arXiv preprint arXiv:1703.02914}, 2017.

\bibitem{li2017gradient}
Yingzhen Li and Richard~E Turner.
\newblock Gradient estimators for implicit models.
\newblock {\em arXiv preprint arXiv:1705.07107}, 2017.

\bibitem{lin2014microsoft}
Tsung-Yi Lin, Michael Maire, Serge Belongie, James Hays, Pietro Perona, Deva
  Ramanan, Piotr Doll{\'a}r, and C~Lawrence Zitnick.
\newblock Microsoft coco: Common objects in context.
\newblock In {\em European conference on computer vision}, pages 740--755.
  Springer, 2014.

\bibitem{liu2016stein}
Qiang Liu and Dilin Wang.
\newblock Stein variational gradient descent: A general purpose {B}ayesian
  inference algorithm.
\newblock In {\em Advances in Neural Information Processing Systems}, pages
  2378--2386, 2016.

\bibitem{louizos2016structured}
Christos Louizos and Max Welling.
\newblock Structured and efficient variational deep learning with matrix
  gaussian posteriors.
\newblock In {\em International Conference on Machine Learning}, pages
  1708--1716, 2016.

\bibitem{louizos2017multiplicative}
Christos Louizos and Max Welling.
\newblock Multiplicative normalizing flows for variational {B}ayesian neural
  networks.
\newblock In {\em International Conference on Machine Learning}, pages
  2218--2227, 2017.

\bibitem{lu2018limitation}
Pei-Hsuan Lu, Pin-Yu Chen, and Chia-Mu Yu.
\newblock On the limitation of local intrinsic dimensionality for
  characterizing the subspaces of adversarial examples.
\newblock {\em arXiv preprint arXiv:1803.09638}, 2018.

\bibitem{ma2018characterizing}
Xingjun Ma, Bo Li, Yisen Wang, Sarah~M Erfani, Sudanthi Wijewickrema, Grant
  Schoenebeck, Dawn Song, Michael~E Houle, and James Bailey.
\newblock Characterizing adversarial subspaces using local intrinsic
  dimensionality.
\newblock In {\em International Conference on Learning Representations (ICLR)},
  2018.

\bibitem{madry2017towards}
Aleksander Madry, Aleksandar Makelov, Ludwig Schmidt, Dimitris Tsipras, and
  Adrian Vladu.
\newblock Towards deep learning models resistant to adversarial attacks.
\newblock {\em arXiv preprint arXiv:1706.06083}, 2017.

\bibitem{madry2018towards}
Aleksander Madry, Aleksandar Makelov, Ludwig Schmidt, Dimitris Tsipras, and
  Adrian Vladu.
\newblock Towards deep learning models resistant to adversarial attacks.
\newblock In {\em International Conference on Learning Representations (ICLR)},
  2018.

\bibitem{mandt2017stochastic}
Stephan Mandt, Matthew~D Hoffman, and David~M Blei.
\newblock Stochastic gradient descent as approximate bayesian inference.
\newblock {\em The Journal of Machine Learning Research}, 18(1):4873--4907,
  2017.

\bibitem{metzen2017detecting}
Jan~Hendrik Metzen, Tim Genewein, Volker Fischer, and Bastian Bischoff.
\newblock On detecting adversarial perturbations.
\newblock In {\em International Conference on Learning Representations (ICLR)},
  2017.

\bibitem{moschoglou2017agedb}
Stylianos Moschoglou, Athanasios Papaioannou, Christos Sagonas, Jiankang Deng,
  Irene Kotsia, and Stefanos Zafeiriou.
\newblock Agedb: the first manually collected, in-the-wild age database.
\newblock In {\em Proceedings of the IEEE Conference on Computer Vision and
  Pattern Recognition Workshops}, pages 51--59, 2017.

\bibitem{osawa2019practical}
Kazuki Osawa, Siddharth Swaroop, Anirudh Jain, Runa Eschenhagen, Richard~E
  Turner, Rio Yokota, and Mohammad~Emtiyaz Khan.
\newblock Practical deep learning with {B}ayesian principles.
\newblock {\em arXiv preprint arXiv:1906.02506}, 2019.

\bibitem{pang2018towards}
Tianyu Pang, Chao Du, Yinpeng Dong, and Jun Zhu.
\newblock Towards robust detection of adversarial examples.
\newblock In {\em Advances in Neural Information Processing Systems (NeurIPS)},
  pages 4579--4589, 2018.

\bibitem{pawlowski2017implicit}
Nick Pawlowski, Andrew Brock, Matthew~CH Lee, Martin Rajchl, and Ben Glocker.
\newblock Implicit weight uncertainty in neural networks.
\newblock {\em arXiv preprint arXiv:1711.01297}, 2017.

\bibitem{rawat2017adversarial}
Ambrish Rawat, Martin Wistuba, and Maria-Irina Nicolae.
\newblock Adversarial phenomenon in the eyes of bayesian deep learning.
\newblock {\em arXiv preprint arXiv:1711.08244}, 2017.

\bibitem{redmon2016you}
Joseph Redmon, Santosh Divvala, Ross Girshick, and Ali Farhadi.
\newblock You only look once: Unified, real-time object detection.
\newblock In {\em Proceedings of the IEEE conference on computer vision and
  pattern recognition}, pages 779--788, 2016.

\bibitem{sengupta2016frontal}
Soumyadip Sengupta, Jun-Cheng Chen, Carlos Castillo, Vishal~M Patel, Rama
  Chellappa, and David~W Jacobs.
\newblock Frontal to profile face verification in the wild.
\newblock In {\em 2016 IEEE Winter Conference on Applications of Computer
  Vision (WACV)}, pages 1--9. IEEE, 2016.

\bibitem{Sharif2016Accessorize}
Mahmood Sharif, Sruti Bhagavatula, Lujo Bauer, and Michael~K. Reiter.
\newblock Accessorize to a crime: Real and stealthy attacks on state-of-the-art
  face recognition.
\newblock In {\em ACM Sigsac Conference on Computer and Communications
  Security}, pages 1528--1540, 2016.

\bibitem{shi2018spectral}
Jiaxin Shi, Shengyang Sun, and Jun Zhu.
\newblock A spectral approach to gradient estimation for implicit
  distributions.
\newblock {\em arXiv preprint arXiv:1806.02925}, 2018.

\bibitem{smith2018understanding}
Lewis Smith and Yarin Gal.
\newblock Understanding measures of uncertainty for adversarial example
  detection.
\newblock {\em arXiv preprint arXiv:1803.08533}, 2018.

\bibitem{sun2017learning}
Shengyang Sun, Changyou Chen, and Lawrence Carin.
\newblock Learning structured weight uncertainty in {B}ayesian neural networks.
\newblock In {\em International Conference on Artificial Intelligence and
  Statistics}, pages 1283--1292, 2017.

\bibitem{sun2018functional}
Shengyang Sun, Guodong Zhang, Jiaxin Shi, and Roger Grosse.
\newblock {Functional} {variational} {{B}ayesian} {neural} {networks}.
\newblock In {\em International Conference on Learning Representations}, 2019.

\bibitem{szegedy2013intriguing}
Christian Szegedy, Wojciech Zaremba, Ilya Sutskever, Joan Bruna, Dumitru Erhan,
  Ian Goodfellow, and Rob Fergus.
\newblock Intriguing properties of neural networks.
\newblock In {\em International Conference on Learning Representations (ICLR)},
  2014.

\bibitem{uesato2018adversarial}
Jonathan Uesato, Brendan O'Donoghue, Aaron van~den Oord, and Pushmeet Kohli.
\newblock Adversarial risk and the dangers of evaluating against weak attacks.
\newblock In {\em International Conference on Machine Learning (ICML)}, 2018.

\bibitem{wang2018cosface}
Hao Wang, Yitong Wang, Zheng Zhou, Xing Ji, Zhifeng Li, Dihong Gong, Jingchao
  Zhou, and Wei Liu.
\newblock Cosface: Large margin cosine loss for deep face recognition.
\newblock In {\em CVPR}, 2018.

\bibitem{welling2011bayesian}
Max Welling and Yee~W Teh.
\newblock Bayesian learning via stochastic gradient langevin dynamics.
\newblock In {\em Proceedings of the 28th international conference on machine
  learning (ICML-11)}, pages 681--688, 2011.

\bibitem{wenzel2020good}
Florian Wenzel, Kevin Roth, Bastiaan~S Veeling, Jakub Swiatkowski, Linh Tran,
  Stephan Mandt, Jasper Snoek, Tim Salimans, Rodolphe Jenatton, and Sebastian
  Nowozin.
\newblock How good is the bayes posterior in deep neural networks really?
\newblock {\em arXiv preprint arXiv:2002.02405}, 2020.

\bibitem{wilson2020bayesian}
Andrew~Gordon Wilson and Pavel Izmailov.
\newblock Bayesian deep learning and a probabilistic perspective of
  generalization.
\newblock {\em arXiv preprint arXiv:2002.08791}, 2020.

\bibitem{xie2017adversarial}
Cihang Xie, Jianyu Wang, Zhishuai Zhang, Yuyin Zhou, Lingxi Xie, and Alan
  Yuille.
\newblock Adversarial examples for semantic segmentation and object detection.
\newblock In {\em Proceedings of the IEEE International Conference on Computer
  Vision}, pages 1369--1378, 2017.

\bibitem{xie2019improving}
Cihang Xie, Zhishuai Zhang, Yuyin Zhou, Song Bai, Jianyu Wang, Zhou Ren, and
  Alan~L Yuille.
\newblock Improving transferability of adversarial examples with input
  diversity.
\newblock In {\em Proceedings of the IEEE Conference on Computer Vision and
  Pattern Recognition (CVPR)}, 2019.

\bibitem{xu2017feature}
Weilin Xu, David Evans, and Yanjun Qi.
\newblock Feature squeezing: Detecting adversarial examples in deep neural
  networks.
\newblock {\em arXiv preprint arXiv:1704.01155}, 2017.

\bibitem{yi2014learning}
Dong Yi, Zhen Lei, Shengcai Liao, and Stan~Z Li.
\newblock Learning face representation from scratch.
\newblock {\em arXiv preprint arXiv:1411.7923}, 2014.

\bibitem{yolov5}
yolov5, 2020.
\newblock \url{https://github.com/ultralytics/yolov5.} Accessed: 2020-05-27.

\bibitem{zhang2018detecting}
Chiliang Zhang, Zuochang Ye, Yan Wang, and Zhimou Yang.
\newblock Detecting adversarial perturbations with saliency.
\newblock In {\em 2018 IEEE 3rd International Conference on Signal and Image
  Processing (ICSIP)}, pages 271--275. IEEE, 2018.

\bibitem{zhang2019theoretically}
Hongyang Zhang, Yaodong Yu, Jiantao Jiao, Eric~P Xing, Laurent~El Ghaoui, and
  Michael~I Jordan.
\newblock Theoretically principled trade-off between robustness and accuracy.
\newblock In {\em International Conference on Machine Learning (ICML)}, 2019.

\bibitem{zheng2018cross}
Tianyue Zheng and Weihong Deng.
\newblock Cross-pose lfw: A database for studying cross-pose face recognition
  in unconstrained environments.
\newblock {\em Beijing University of Posts and Telecommunications, Tech. Rep},
  5, 2018.

\bibitem{zheng2017cross}
Tianyue Zheng, Weihong Deng, and Jiani Hu.
\newblock Cross-age lfw: A database for studying cross-age face recognition in
  unconstrained environments.
\newblock {\em arXiv preprint arXiv:1708.08197}, 2017.

\end{thebibliography}
}

\clearpage
\appendix
\section{Attack Methods}
\label{sec:attack}
In this part, we outline the details of the adopted attack methods in this paper. 
For simplicity, we use $l(\bm{x},y)$ to notate the loss function for attack which inherently connects to the negative log data likelihood, e.g., the cross entropy in image classification, the pairwise feature distance in open-set face recognition, and the weighted sum of the bounding-box regression loss and the classification loss in object detection.

\textbf{FGSM} \cite{goodfellow2014explaining} crafts an adversarial example under the $\ell_{\infty}$ norm as 
\begin{equation}
    \bm{x}^{\text{adv}} = \bm{x} + \epsilon\cdot\mathrm{sign}(\nabla_{\bm{x}}l(\bm{x},y))\text{,}
\end{equation}
FGSM can be extended to an $\ell_{2}$ attack as

\begin{equation}
    \bm{x}^{\text{adv}} = \bm{x} + \epsilon\cdot\frac{\nabla_{\bm{x}}l(\bm{x},y)}{\|\nabla_{\bm{x}}l(\bm{x},y)\|_2}\text{.}
\end{equation}
In all experiments, we set the perturbation budget $\epsilon$ as $16/255$.

\textbf{BIM} \cite{kurakin2016adversarial} extends FGSM by taking iterative gradient updates:
\begin{equation}
\label{eq:bim}
    \bm{x}_{t+1}^{\text{adv}} = \mathrm{clip}_{\bm{x},\epsilon} \big(\bm{x}_t^{\text{adv}} + \eta\cdot\mathrm{sign}(\nabla_{\bm{x}}l(\bm{x}_t^{\text{adv}},y))\big),
\end{equation}
where $\mathrm{clip}_{\bm{x},\epsilon}$ guarantees the adversarial example to satisfy the $\ell_{\infty}$ constraint. 
For all the iterative attack methods, we set the number of iterations as $20$ and the step size $\eta$ as $1/255$.

\textbf{PGD}~\cite{madry2018towards} complements BIM with a random initialization for the adversarial examples (i.e., $\bm{x}_0^{\text{adv}}$ is uniformly sampled from the neighborhood of $\vx$).

\textbf{MIM} \cite{Dong2017} introduces a momentum term into BIM as 
\begin{equation}
    \bm{g}_{t+1} = \mu \cdot \bm{g}_t + \frac{\nabla_{\bm{x}}l(\bm{x}_t^{\text{adv}},y)}{\|\nabla_{\bm{x}}l(\bm{x}_t^{\text{adv}},y)\|_1}\text{,}
\end{equation}
where $\mu$ refers to the decay factor and is set as 1 in all experiments.
Then, the adversarial example is calculated by
\begin{equation}
    \bm{x}^{\text{adv}}_{t+1}=\mathrm{clip}_{\bm{x},\epsilon}(\bm{x}^{\text{adv}}_t+\eta\cdot\mathrm{sign}(\bm{g}_{t+1}))\text{.}
\end{equation}
We adopt the same hyper-parameters as in BIM.

\textbf{DIM}~\cite{xie2019improving} relies on a stochastic transformation function to craft adversarial examples, which can be represented as
\begin{equation}
\label{eq:dim}
    \bm{x}_{t+1}^{\text{adv}} = \mathrm{clip}_{\bm{x},\epsilon} \big(\bm{x}_t^{\text{adv}} + \eta\cdot\mathrm{sign}(\nabla_{\bm{x}}l(T(\bm{x}_t^{\text{adv}}; p), y))\big),
\end{equation}
where $T(\bm{x}_t^{\text{adv}}; p)$ refers to some transformation to diversify the input with probability $p$.

\textbf{TIM}~\cite{dong2019evading} integrates the translation-invariant method into BIM by convolving the gradient with the pre-defined kernel $\bm{W}$ as
\begin{equation}
\label{eq:tim}
    \bm{x}_{t+1}^{\text{adv}} = \mathrm{clip}_{\bm{x},\epsilon} \big(\bm{x}_t^{\text{adv}} + \eta\cdot\mathrm{sign}(\bm{W} * \nabla_{\bm{x}}l(\bm{x}_t^{\text{adv}}, y))\big).
\end{equation}

\textbf{C\&W} adopts the original C\&W loss~\cite{carlini2017towards} based on the iterative mechanism of BIM to perform attack in classification tasks. 
In particular, the loss takes the form of
\begin{equation}
\label{eq:cw}
     l_{cw} = \max(Z(\bm{x}^{\text{adv}}_{t})_y - \max_{i\neq y}Z(\bm{x}^{\text{adv}}_{t})_i, 0) \text{,}
\end{equation}
where $Z(\bm{x}^{\text{adv}}_{t})$ is the logit output of the classifier.

\textbf{SPSA} \cite{uesato2018adversarial} estimates the gradients by
\begin{equation}
    \bm{\hat{g}} = \frac{1}{q}\sum_{i=1}^{q}\frac{l(\bm{x}+\sigma \bm{u}_i,y)-l(\bm{x}-\sigma\bm{u}_i,y)}{2\sigma}\cdot \bm{u}_i,
\end{equation}
where $\{\bm{u}_i\}_{i=1}^q$ are samples from a Rademacher distribution, and we set $\sigma=0.001$ and $q=64$. Besides, $l(\bm{x}, y) = Z(\bm{x})_y - \max_{i\neq y}Z(\bm{x})_i$ is used in our experiments rather than the cross entropy loss.
We take an Adam~\cite{kingma2014adam} optimizer with 0.01 learning rate to apply the estimated gradients.

\section{More Experiment Details}
For $\ell_2$ threat model, we adopt the normalized $\ell_2$ distance  $\bar{\ell}_2(\bm{a}) = \frac{\|\bm{a}\|_2}{\sqrt{d}}$ as the measurement, 
where $d$ is the dimension of a vector $\bm{a}$.
The decay factors of MIM, TIM, and DIM are $1.0$.

In ImageNet classification, we apply Gaussian blur upon the sampled uniform noise with 0.03 probability, and then use the outcome to perturb the training data.
The technique can enrich the training perturbations with low-frequency patterns, promoting the adversarial detection sensitiveness against diverse kinds of adversarial perturbations.

To attack the open-set face recognition system in the evaluation phase, we find every face pair belonging to the same person, and use one of the paired faces as $\vx$ and the feature of the other as $y$ to perform attack.
The loss function for such an attack is the $\ell_2$ distance between $y$ and the feature of $\vx$ (as mentioned in Sec~\ref{sec:attack}).
As the \emph{posterior predictive} is not useful in such an open-set scenario, we perform \emph{Bayes ensemble} on the output features and then leverage the outcomes to make decision.
Due to the limited GPU memory, we attack the deterministic features of the \emph{MC dropout} baseline instead of the features from \emph{Bayes ensemble}, while the uncertainty estimates are still estimated based on 20 stochastic forward passes with dropout enabled.

In object detection, we adopt the YOLOV5-s architecture, there are three feature output heads (\texttt{BottleneckCSP} modules) to deliver features in various scales.
Thus, we make these three heads be Bayesian when implementing \emph{LiBRe}.
During inference, we average the features calculated given different parameter candidates to obtain an assembled feature to detect objects, which assists us to bypass the potential difficulties of directly assembling the object detection results.

\section{Generalization to Score-based Attack} 
We additionally concern whether \emph{LiBRe} can generalize to the adversarial examples generated by score-based attacks, which usually present different characteristics from the gradient-based ones.
We leverage the typical SPSA~\cite{uesato2018adversarial} to conduct experiments on ImageNet, getting 0.969 detection AUROC. 
This further evidences our \emph{attack agnostic} designs.

\section{Detect More Ideal Attacks}
At last, we evaluate the adversarial detection ability of \emph{LiBRe} on more ideal attacks.
We add the constraint that the generated adversarial examples should also have small predictive uncertainty into the existing attacks. 
This means that the attacks can jointly fool the decision making and uncertainty quantification aspects of the model.
We add an uncertainty minimization term upon the original attack objective to implement this.
We feed the crafted adversarial examples into \emph{LiBRe} to assess if they can be identified.
On ImageNet, we obtain the following adversarial detection AUROCs: 0.9996, 0.2374, 0.0363, 0.2211, 0.1627, 0.1990, 0.9998, 0.9627, 0.2537, and 0.2213 for
FGSM, BIM, C\&W, PGD, MIM, TIM, DIM, FGSM-$\ell_2$, BIM-$\ell_2$, and PGD-$\ell_2$, respectively.

The results reveal that \emph{LiBRe} is likely to be defeated if being fully exposed to the attackers.
But it is also no doubt that \emph{LiBRe} is powerful enough if keeping opaque to the attack algorithms as the pioneering work~\cite{feinman2017detecting}.
We believe that introducing adversarial training mechanism into \emph{LiBRe} would significantly boost the ability of detecting these ideal attacks, and we leave it as future work.

\end{document}